\documentclass[11pt]{article}
\PassOptionsToPackage{table}{xcolor}
\usepackage[preprint]{acl}

\usepackage{times}
\usepackage{latexsym}
\usepackage{xcolor}
\usepackage{booktabs}
\usepackage{graphicx}
\usepackage{subcaption}

\usepackage{amsmath}
\usepackage{xurl}
\usepackage[T1]{fontenc}
\usepackage[utf8]{inputenc}

\usepackage{microtype}

\usepackage{inconsolata}

\usepackage{graphicx}
\usepackage{xcolor}
\usepackage{tcolorbox}
\usepackage{mdframed}
\newtcolorbox{promptbox}[1]{
width=\linewidth,
title={#1},
fonttitle=\bfseries\footnotesize,
fontupper=\ttfamily\scriptsize,
before upper=\raggedright,
before skip=2pt,
after skip=4pt
}
\tcbuselibrary{listings,breakable}

\usepackage{booktabs}
\usepackage{amssymb}
\usepackage{enumitem}
\usepackage{float}
\usepackage{placeins}
\usepackage{needspace}
\usepackage{wrapfig}

\tcbset{
  colback=gray!6!white,
  colframe=gray!40,
  boxrule=0.2pt,
  arc=1pt,
  left=2pt,
  right=2pt,
  top=1pt,
  bottom=1pt
}

\title{FactReview: Evidence-Grounded Peer Review with Execution-Based Claim Verification}

\author{
  \textbf{Ling Yue\textsuperscript{1,*}},
  \textbf{Chaoqian Ouyang\textsuperscript{2,*}},
  \textbf{Ruijun Huang\textsuperscript{3,*}},
  \textbf{Hang Xu\textsuperscript{3}}, 
  \textbf{Lu Li\textsuperscript{2}}, \\
  \textbf{Yuchen Liu\textsuperscript{4}}, 
  \textbf{Libin Zheng\textsuperscript{2}},
  \textbf{Wei Liu\textsuperscript{2}}, 
  \textbf{Shaowu Pan\textsuperscript{1,\textdagger}},
  \textbf{Shimin Di\textsuperscript{3,\textdagger}},
  \textbf{Min-Ling Zhang\textsuperscript{3}} \\
  \textsuperscript{1}Rensselaer Polytechnic Institute, 
  \textsuperscript{2}Sun Yat-sen University,
  \textsuperscript{3}Southeast University \\
  \textsuperscript{4}The Hong Kong University of Science and Technology
}

\begin{document}

\maketitle
\renewcommand{\thefootnote}{\fnsymbol{footnote}}
\footnotetext[1]{Equal contribution.}
\footnotetext[2]{\raggedright Corresponding authors:
\texttt{shimin.di@seu.edu.cn};
\texttt{pans2@rpi.edu}\par}
\renewcommand{\thefootnote}{\arabic{footnote}}
\begin{abstract}
AI is accelerating scientific research, but providing timely and trustworthy feedback remains challenging. Such feedback requires evaluating specific claims against verified evidence, 
helping researchers and autonomous research systems determine which findings can guide subsequent work. Thorough verification is time-consuming, and reviewers under time pressure may overlook or misjudge critical issues. Existing AI reviewers have yet to fill this gap because they often assess claims without adequately verifying the supporting evidence. We introduce FactReview, a framework that verifies claims in papers to provide more trustworthy feedback. It checks claims against the literature and examines derivations, while also executing the paper’s code to test empirical claims. Across 300 papers, FactReview ranks best among all evaluated systems on five of seven metrics, achieving 6.8\% false support and 79.2\% classification accuracy. In a study of 36 papers, reviewers using FactReview detect 9.4 issues per paper compared with 5.6 independently, while spending 74.1\% less time. FactReview thus supports more reliable peer review, with the potential to accelerate research iteration. The code is public at {\url{https://github.com/DEFENSE-SEU/FactReview}}.

\end{abstract}
\begin{table*}[t]
\centering

{
\begin{tabular*}{\textwidth}{@{\extracolsep{\fill}}lccccc@{}}
\toprule

\raisebox{4pt}[0pt][0pt]{\textbf{System}}
& \shortstack{\textbf{Retrieved}\\\textbf{literature}}
& \shortstack{\textbf{Claim}\\\textbf{assessment}}
& \shortstack{\textbf{Evidence-linked}\\\textbf{review}}
& \shortstack{\textbf{Code}\\\textbf{execution}}
& \shortstack{\textbf{No final}\\\textbf{recommendation}} \\
\midrule
MARG             & -- & $\checkmark$ & $\triangle$ & -- &$\checkmark$\\
OpenReviewer     & -- & $\checkmark$ & $\triangle$ & -- & -- \\
AgentReview      & -- & $\triangle$ & -- & -- & -- \\
ReviewerToo      & $\checkmark$ & $\checkmark$ & $\checkmark$ & -- & -- \\
DeepReviewer 2.0 & $\checkmark$ & -- & $\checkmark$ & -- & -- \\
PaperBench       & -- & -- & -- & $\checkmark$ & $\checkmark$ \\
\midrule
\textbf{FactReview} & $\checkmark$ & $\checkmark$ & $\checkmark$ & $\triangle$ & $\checkmark$\\
\bottomrule

\end{tabular*}
}
\caption{$\checkmark$ explicit; $\triangle$ partial or conditional;
``--'' outside the reported task. Execution includes experiments or released
code; decision output is a score or recommendation.}
\label{tab:system_comparison}
\end{table*}
\section{Introduction}

AI systems are accelerating scientific research by automating tasks across the research process, from developing ideas to writing manuscripts \citep{lu2024ai,tang2026ai}. This acceleration increase the risk of producing work whose quality cannot be guaranteed, since such claims are often difficult to verify. Without trustworthy feedback to identify incorrect claims, researchers and autonomous systems risk carrying these errors into subsequent research or abandoning promising directions based on unfounded criticism.
Providing such feedback requires tracing the underlying reasoning and running code to check whether the reported results support the conclusions. However, human reviewers have limited time to check the evidence behind each claim, and growing research output adds to this pressure, leading to rushed assessments and increasing the risk of errors \citep{liang2024can,neurips2025reviewprocess}. These constraints create a need for automated reviewers that investigate suspected errors and support their judgments with evidence.

\begin{figure}[!t]
    \centering
    \includegraphics[width=1.0\linewidth]{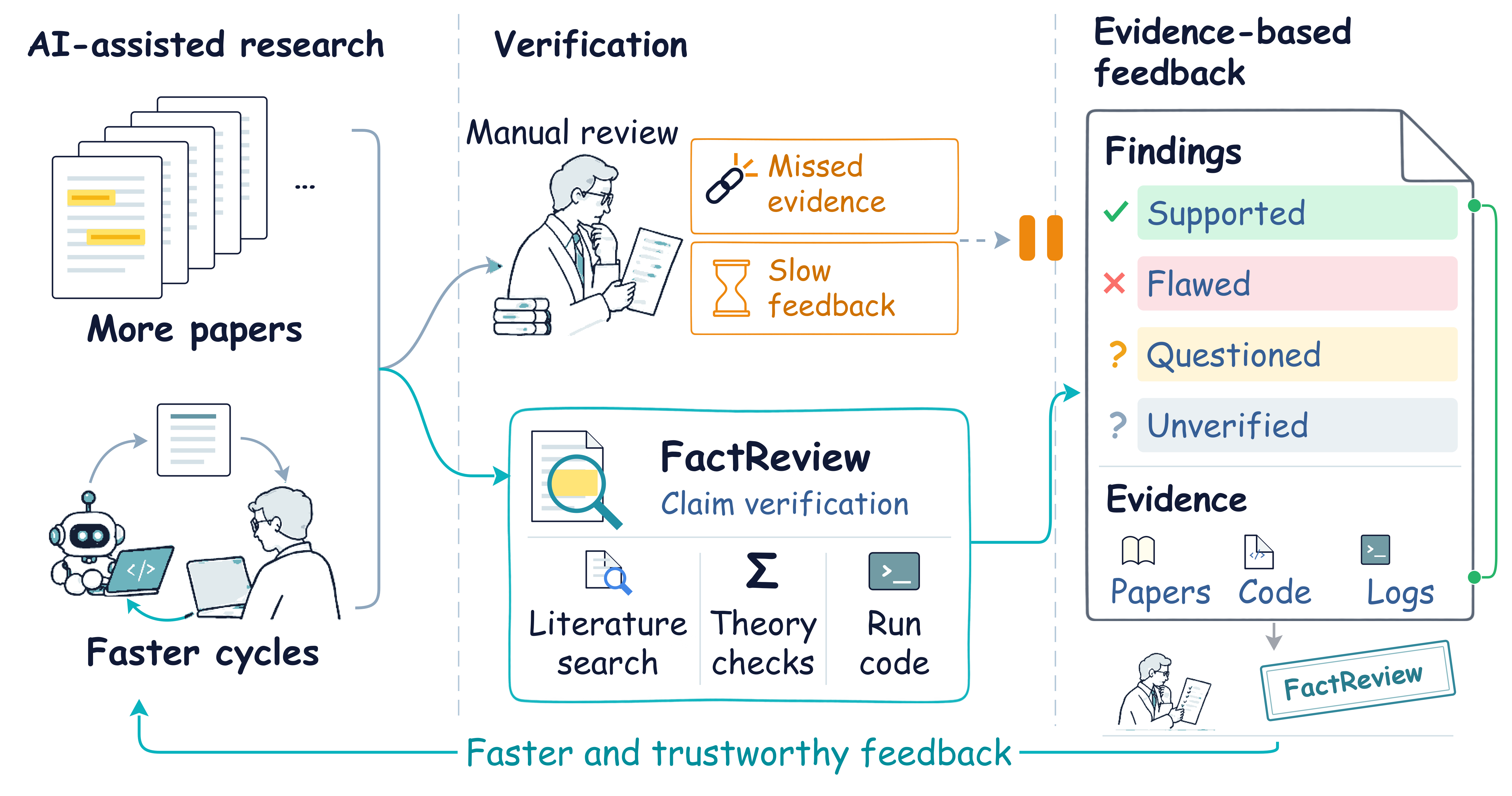} 
    \caption{Motivation of FactReview. Faster research demands timely, trustworthy feedback. FactReview verifies claims against evidence before generating reviews, helping researchers and AI scientists determine which findings can reliably guide subsequent work.}
    \label{fig:motivation}
\end{figure}


Existing AI reviewers have yet to meet this demand for trustworthy feedback. They can generate reviews at scale \citep{d2024marg,idahl2025openreviewer}, but recent systems \citep{sahu2025reviewertoo,zhu2025deepreview,garg2025revieweval} often assess claims without adequately verifying the supporting evidence \citep{dycke2026automatic}. An AI reviewer may accept an experimental claim without checking whether the paper's code produces results that support it \citep{jiang2026badscientist}. Table~\ref{tab:system_comparison} compares these systems with FactReview, highlighting the gap between claim assessment and code execution in prior systems. Researchers must therefore verify these judgments themselves before acting on the feedback they receive.

We introduce FactReview, a framework that verifies claims in research papers to provide trustworthy feedback. It checks claims against relevant literature and examines theoretical derivations. For empirical claims, FactReview executes the paper's code to test whether the results support those claims. Each judgment links a claim in the paper to the evidence used to assess it.


Evaluation on 300 papers shows that FactReview ranks best among all evaluated systems on five of seven metrics. It achieves the lowest false support rate at 6.8\% and the highest claim classification accuracy at 79.2\%. It also leads in issue detection, with recall of 76.4\% overall and 82.1\% for core claims. Ablation results show that each verification branch contributes to issue detection, with literature verification improving recall by 14.6 percentage points. In a study involving 36 papers, reviewers using FactReview identify an average of 9.4 issues per paper, compared with 5.6 during independent review, while spending 74.1\% less time. These results demonstrate that FactReview supports faster, more trustworthy peer review, with the potential to accelerate research iteration.

Our main contributions are as follows:
\begin{itemize}[leftmargin=*]
    \item We redefine AI review as fact-grounded claim verification, providing traceable feedback that supports reviewers, researchers, and AI Scientists rather than replacing them.
    \item FactReview supports more trustworthy AI review. It checks claims against the literature and examines derivations, while also executing the paper's code to test empirical claims.
    \item Controlled evaluation shows that FactReview outperforms all evaluated baselines on five of seven metrics. In a reviewer-assistance study, its reports reduce average review time by  74.1\% and increase the average number of issues detected per paper from 5.6 to 9.4.
\end{itemize}

\section{Related Work}
\paragraph{Autonomous research and AI reviewing.}
Autonomous research systems accelerate scientific workflows \citep{lu2024ai,tang2026ai}. Agent Laboratory incorporates human feedback \citep{schmidgall2025agent}, while AI co-scientist and CycleResearcher use iterative feedback to guide research \citep{gottweis2025towards,weng2025cycleresearcher}. AgentRxiv enables agents to build on shared results \citep{schmidgall2025agentrxiv}. This reuse creates a need for reviews that detect errors before they influence subsequent studies. LLM-based reviewing offers a way to scale such feedback \citep{zhang2024comprehensive}, with systems developed for automated review generation \citep{gao2024reviewer2,sahu2025reviewertoo}. MARG refines feedback through agent discussions \citep{d2024marg}. OpenReviewer uses fine-tuning on expert reviews \citep{idahl2025openreviewer}, while DeepReview incorporates literature retrieval \citep{zhu2025deepreview}. Detecting errors, however, can require evidence from code execution that is absent from the manuscript \citep{luo2025more}. Reproducibility studies underscore the importance of inspecting research artifacts \citep{raff2019step,pineau2021improving} and the difficulty of executing them reliably. CLAIMCHECK audits whether completed LLM critiques are grounded in the manuscript claims they discuss \citep{ou2025claimcheck}. FactReview establishes evidence before generating the review by combining external reference checks with code execution (Table~\ref{tab:system_comparison}). It connects this evidence to the specific claims under review, making its judgments auditable before they guide further research.
\begin{figure*}[t]
\centering
\includegraphics[width=0.92\textwidth]{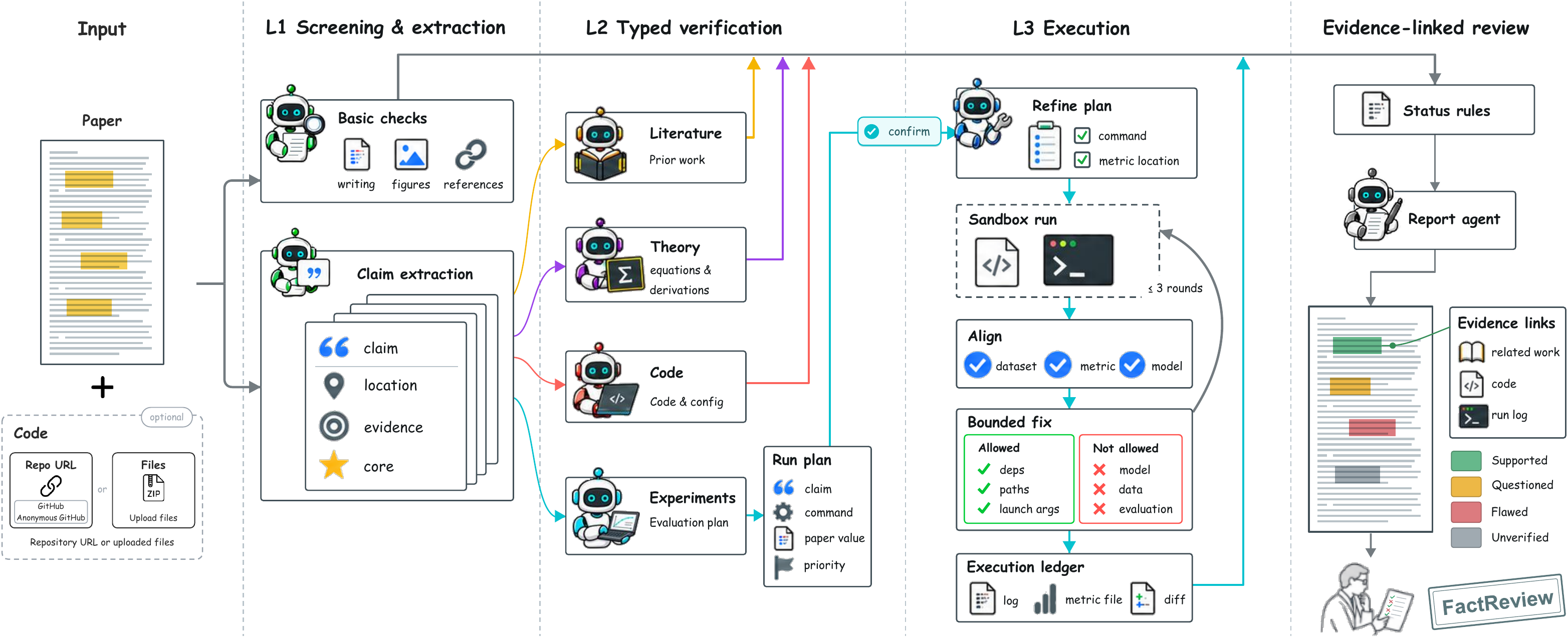}
\caption{FactReview extracts review-relevant claims, constructs manuscript, literature, reference-integrity, and optional execution evidence, assigns conservative outcomes, and links the review to evidence.}
\label{fig:overview}
\end{figure*}


\paragraph{Evidence-grounded scientific assistance.}
Scientific claim verification asks whether retrieved evidence supports or refutes a claim \citep{wadden2020fact,soleimani2020bert}. Literature-grounded assistants such as PaperQA~\citep{lala2023paperqa}, OpenScholar~\citep{asai2024openscholar}, and SF-RAG~\citep{yu2026sf} answer scientific questions with stronger provenance than paper-only prompting, using scholarly-document infrastructure such as SciBERT~\citep{beltagy2019scibert}, S2ORC~\citep{lo2020s2orc}, and GROBID~\citep{grobid}. Work on reproducibility and execution-based evaluation shows that released artifacts often contain evidence absent from the manuscript \citep{raff2019step,pineau2021improving}, while agentic reproduction benchmarks show that bridging paper, code, environment, and reported results remains difficult \citep{starace2025paperbench,zhao2026autoreproduce,hua2026researchcodebench}. These efforts provide useful building blocks, but they do not produce reviewer-facing claim assessments.

\paragraph{Execution and reproducibility agents.}
PaperBench \citep{starace2025paperbench} evaluates agents' ability to reproduce research results. AutoReproduce \citep{zhao2026autoreproduce} draws on cited papers to guide reproduction, while CORE-Bench \citep{siegel2024core} tests reproduction using authors' code and data. ResearchCodeBench \citep{hua2026researchcodebench} evaluates whether models can implement methods described in papers. These methods evaluate implementation or reproduction success, while FactReview determines how execution evidence supports or challenges individual paper claims. It checks that the executed experiment matches the claim's stated conditions before using the result to support a review judgment.

\section{Method}\label{sec:method}

\subsection{Overview}\label{sec:overview}

Given a submitted paper and its public code when available, FactReview verifies the claims in the paper one at a time and collects, for each claim, evidence that a reviewer can inspect directly. Each verification result is a tuple
\begin{equation}
r_i = (c_i,\ E_i,\ s_i,\ a_i).
\label{eq:result}
\end{equation}
The claim $c_i$ carries its page number and the passage it is taken from. Each piece of evidence in $E_i$ records its source, which may be a passage of the paper, a retrieved publication, a location in the code, or a run log. The status $s_i$ is one of Supported, Flawed, Questioned and Unverified. The advice $a_i$ tells the reviewer how to use the result. It may state a problem the reviewer can raise directly, a doubt to put to the authors, or the reason the claim could not be verified. We attach every status to a specific claim and its evidence so that the result can be trusted. For any conclusion, the reviewer can see which sentence it concerns and what it rests on.

Before verification begins, FactReview converts the paper and the code into material shared by all modules. MinerU parses the paper into sectioned text and images that keep their page positions, and the repository is indexed by file. Each module reads only the sections, figures or code files it needs.

As shown in Figure~\ref{fig:overview}, verification proceeds in three layers of increasing depth. Layer~1 works from the paper alone. The claims it extracts are checked in Layer~2 against the literature, derivations, code and experimental design, and the execution plans written in Layer~2 are carried out in Layer~3 to obtain new evidence by running the code. The evidence from all three layers is then pooled to decide the status of each claim (Section~\ref{sec:status}).

\subsection{Layered Verification}\label{sec:layers}

\noindent\textbf{Layer 1: surface checks and claim extraction.} Layer~1 checks the writing, the figures and tables, and the references, and it extracts the claims that later layers verify.

In the writing check, an LLM reads the paper section by section and quotes each sentence that has a problem. It looks for language errors such as typos and unclear phrasing, cross-references to figures, tables or sections that point nowhere, and breaches of submission rules such as anonymity. In the figure check, a vision-language model reads each figure and table together with its caption and the sentences that cite it. It judges whether the figure is self-contained, legible and consistent with the text, for example whether the axes are labeled and whether bold entries in a table are explained. The reference check matches each entry against records in scholarly databases. An entry with no matching paper is flagged as possibly fabricated. An entry that matches a paper but gives the wrong authors, year or venue is flagged as incorrect and receives a corrected version. Retracted papers are flagged separately.

The claims of a paper are spread across its sections, from the contributions stated in the introduction to the design choices in the method and the results in the experiments. For claim extraction, an LLM reads the full text and lists every claim that the authors state explicitly and that can be verified on its own. A statement whose parts need different evidence or could reach different verdicts is split into several claims. Each claim keeps its source text, and the system confirms that this text appears in the paper. Each claim is finally tagged with the kinds of evidence it requires and with whether it is a core claim, and it is passed to the matching branches of Layer~2.

\noindent\textbf{Layer 2: verification by evidence type.} Different claims need different evidence. Novelty is judged against prior work and a theorem against its derivation. Implementation details are checked against the code, and an experimental conclusion depends on whether the experiment behind it is well designed. Layer~2 therefore consists of four agents, each responsible for one kind of evidence.

The literature agent searches for related work that the paper does not cite and selects the closest papers. It compares them with the submission in method, problem setting and evaluation to test whether the claimed novelty holds. It also checks whether each cited paper supports the sentence that cites it, and it points out important work and baselines that the paper omits. The theory agent redoes the derivations in the main text step by step and consults the proofs in the appendix when needed, looking mainly for skipped steps, unstated assumptions and inconsistent notation. The code agent uses the repository index to check whether the model, training and evaluation described in the paper match the code. The experiment agent works from the experimental section and checks that each experimental conclusion has a corresponding experiment, that comparisons are fair, and that key components are ablated to show their individual effects.

Every piece of evidence from these agents points to a specific location, such as a paragraph of the paper, a retrieved publication or a line of code, and states whether it supports the claim or reveals a problem. This evidence forms $E_i$ in Eq.~\eqref{eq:result}.
\begin{table*}[t]
\centering\small
\setlength{\tabcolsep}{3pt}
\renewcommand{\arraystretch}{1.12}
\begin{tabular}{lccccccc}
\toprule
System & \shortstack{False support\\$\downarrow$} & \shortstack{Coverage\\$\uparrow$} & \shortstack{Issue recall\\$\uparrow$} & \shortstack{Core issue\\recall $\uparrow$} & \shortstack{Claim F1\\$\uparrow$} & \shortstack{Accuracy\\$\uparrow$} & \shortstack{Reviewer issue\\recall $\uparrow$} \\
\midrule
Direct & 29.6 & 76.8 & 43.7 & 48.2 & 78.6 & 56.9 & 40.5 \\
DeepReviewer 2.0 & 23.4 & 73.5 & 57.8 & 63.1 & 79.8 & 63.2 & 58.7 \\
ReviewGrounder & 20.7 & 75.9 & 61.3 & 66.8 & 78.9 & 65.6 & 62.4 \\
Claude Code & 16.8 & \underline{80.4} & 68.9 & \underline{74.6} & \underline{84.8} & 71.5 & 66.8 \\
Codex & \underline{15.3} & \textbf{81.7} & \underline{70.6} & 73.8 & \textbf{85.1} & \underline{72.8} & \underline{68.3} \\
\midrule
\rowcolor{gray!12}
\textbf{FactReview} & \textbf{6.8} & 77.6 & \textbf{76.4} & \textbf{82.1} & 84.3 & \textbf{79.2} & \textbf{74.6} \\
\bottomrule
\end{tabular}
\caption{Comparison with baseline systems. All values are percentages. Status metrics use annotated claims from all 300 papers. Claim F1 uses the 60 complete inventories. Reviewer issue recall uses the 222 papers with eligible public review issues. Best results are in bold. Second-best results are underlined.}
\label{tab:main_results}
\end{table*}
Some experimental conclusions can only be confirmed by running the code. For each such claim, the experiment agent writes an execution plan and hands it to Layer~3. The plan records the candidate commands found in the repository, the run mode, the values reported in the paper, and whether the plan can currently be run. The run mode is evaluation, analysis or training. A plan that cannot run states why, for example because code, data or model weights are missing. Each plan also receives a priority according to the importance of its claim, and plans for core claims rank higher. Details such as exact command arguments and the file where metrics are written are left for Layer~3 to fill in.

\noindent\textbf{Layer 3: execution.} An execution agent runs the authors' code to add evidence for claims that only a run can confirm. Before it starts, the user confirms the execution plans and their estimated cost. Training is expensive, so the agent trains only for core claims whose plans fit the budget and restricts the remaining plans to evaluation or analysis. Plans run in order of priority, and within a priority level the plans that can run directly go first. Plans that cannot run yet are retained to explain why execution evidence is unavailable.

Before each run, the agent searches the repository to complete what the plan leaves open, such as the full command and the location of the metric output. It checks the documentation, scripts and configuration files to resolve the working directory, required inputs and command arguments. It also identifies the log entry or result file from which the metric will be read. The code runs in an isolated sandbox, and the whole process is logged. The command, logs and metric files remain linked to the plan so that the reviewer can trace the result to its execution.

A run that completes may still fail to test the claim, since its output has to correspond to the claim. Let $o$ denote the result of a run. It counts as evidence for claim $c_i$ only if
\begin{equation}
(d_o,\ m_o,\ \theta_o) = (d_{c_i},\ m_{c_i},\ \theta_{c_i}).
\label{eq:aligned}
\end{equation}
Here $d$ denotes the dataset and its split, $m$ the metric and how it is computed, and $\theta$ the evaluated model. The subscript $c_i$ marks the setting under which the paper reports the result, which the execution plan already specifies.

If a run fails, the agent examines the failed command and its logs to identify the cause. It may correct dependencies or paths and adjust launch arguments while leaving the experiment itself unchanged. Each repair must address an observed failure and preserve the model, data and evaluation. The agent makes at most three repair rounds and records every change with its file difference. After each repair, it reruns the command and checks the recovered result against Eq.~\ref{eq:aligned}. A numerical discrepancy alone does not justify modifying the experiment or its evaluation.
\begin{figure*}[t]
\centering
\begin{subfigure}[b]{0.38\textwidth}
\centering
\includegraphics[width=\linewidth]{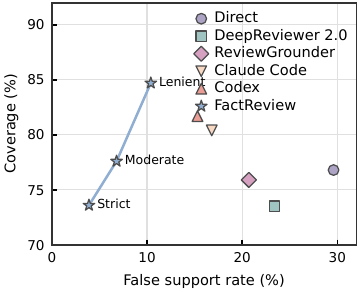}
\caption{Gate levels}
\label{fig:gate_levels}
\end{subfigure}\hfill
\begin{subfigure}[b]{0.54\textwidth}
\centering
\includegraphics[width=\linewidth]{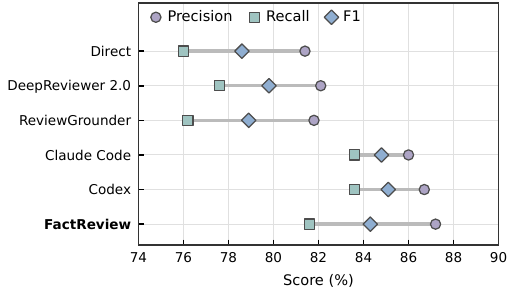}
\caption{Claim extraction}
\label{fig:claim_extraction}
\end{subfigure}
\caption{(a) Coverage and false support rate of the baselines and of FactReview at 3 gate levels. (b) Claim extraction precision, recall, and F1 on the 60 complete inventories.}
\label{fig:gate_extraction}
\end{figure*}
An aligned result is compared with the value reported in the paper and counts as inconsistent when the gap clearly exceeds the variation the paper reports. For an inconsistent result, the agent records whether the discrepancy appears directly in data or logs released by the authors, since such a discrepancy predates the sandbox run. All runs and repairs are compiled into an execution record that the reviewer receives with the report.

\subsection{Status Assignment and Reporting}\label{sec:status}

Once the evidence from all layers is pooled, each claim receives a status. A claim is Supported when the evidence suffices to support it and Unverified when the evidence is too thin for a conclusion. When a problem is found, the status depends on whether the authors could explain it. The claim is Flawed if no reasonable explanation would change the conclusion, and Questioned if a response from the authors could resolve the doubt. A claim is also Questioned when the supporting and opposing evidence are equally strong, and the report then lists both sides for the reviewer to weigh. An LLM judges whether each piece of evidence is sufficient and which side it favors, and a program assigns the final status by fixed rules.

After the status is set, a report agent writes the advice $a_i$ for each claim. For a Flawed claim, the advice states the problem the reviewer can raise directly. A Questioned claim receives advice phrased as a question for the authors, and the advice for an Unverified claim names the materials that are missing. The report opens with a summary of the overall results that highlights the most important problems. It then lists the claims in the order Flawed, Questioned, Unverified and Supported, and each claim can be traced to its source text and evidence. The writing, figure and reference problems found in Layer~1 follow, and the execution record comes last. The report makes no recommendation to accept or reject the paper and leaves that decision entirely to the human reviewer.

\section{Experiments}
\label{sec:experiments}

\subsection{Experimental Setup}

\noindent\textbf{Dataset.}
The dataset contains 300 papers across 6 groups. AI generated papers, ICLR accepted papers, ICLR rejected papers, and TMLR papers each contribute 52, Nature Communications contributes 26, and 66 recent papers are selected by first public version date to examine performance on newer research. Sampling is stratified by source and weighted by reviewer issue density, with code availability also considered. Of the 300 papers, 60 have complete claim inventories, and the remaining 240 are annotated for selected key claims. Annotators check claims extracted by FactReview's preparation pipeline against the full manuscript, remove invalid entries, and add omitted claims before assigning statuses. On 500 double annotated claims, Cohen's $\kappa$ is 0.757. Dataset construction and annotation guidelines are provided in Appendix~\ref{app:dataset}.

\noindent\textbf{Baselines.}
FactReview is compared with 5 baselines. Direct extracts and assesses claims in a single model call using the paper and available code documentation. DeepReviewer 2.0 retrieves related literature and annotates the paper before generating a review. ReviewGrounder revises an initial review through analysis of the paper's contributions and experiments. Claude Code and Codex receive the same verification instructions as FactReview and can gather evidence or run code within the allocated budget. Reports from the dedicated reviewing systems are converted into claim assessments, and claims absent from a report remain Unverified.
\begin{figure*}[t]
\centering
\begin{subfigure}[b]{0.31\textwidth}
\centering
\includegraphics[width=\linewidth]{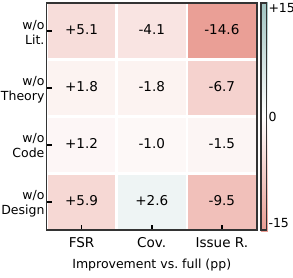}
\caption{Verification branches}
\label{fig:ablation_branches}
\end{subfigure}\hfill
\begin{subfigure}[b]{0.31\textwidth}
\centering
\includegraphics[width=\linewidth]{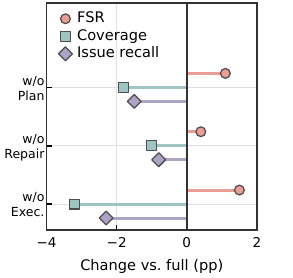}
\caption{Execution module}
\label{fig:ablation_execution}
\end{subfigure}\hfill
\begin{subfigure}[b]{0.31\textwidth}
\centering
\includegraphics[width=\linewidth]{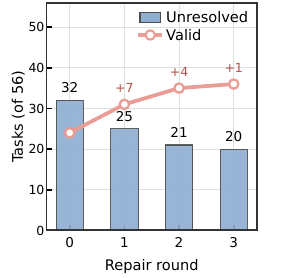}
\caption{Repair rounds}
\label{fig:repair_rounds}
\end{subfigure}

\caption{(a) Change from the full system when removing each verification branch, on all 300 papers. (b) The same for execution components, on the 28 papers with code. (c) Unresolved and valid tasks among 56 planned tasks after each of the 3 repair rounds, starting from no repair.}
\label{fig:ablation}
\end{figure*}
\noindent\textbf{Metrics.}
False support rate measures the proportion of reference Flawed or Questioned claims labeled Supported. Coverage measures the proportion of reference Supported claims also labeled Supported. Accuracy requires an exact match across the 4 statuses. Issue recall measures the recovery of reference Flawed or Questioned claims, with core issue recall restricted to claims supporting the main contribution. Reviewer issue recall uses 222 papers with eligible concerns from public reviews, and claim extraction F1 uses the 60 complete inventories. Definitions are provided in Appendix~\ref{app:metrics}.

\noindent\textbf{Implementation Details.}
Claude Code uses Claude Opus 5.0, and the remaining systems use gpt-5.5. All systems receive a total budget of 60 minutes and 10 million tokens per paper, including code execution. Repairs adjust the environment and configuration and leave the scientific implementation unchanged. Execution settings and cost are provided in Appendix~\ref{app:implementation}.

\subsection{Main Results}

Table~\ref{tab:main_results} shows that FactReview lowers the rate of false support while keeping its coverage close to that of the general purpose agents. Its false support rate is 6.8\%, compared with 15.3\% for Codex, and its coverage is 77.6\%, compared with 81.7\% for Codex. A paired bootstrap analysis over papers places the reduction in false support at 8.5 percentage points, with a 95\% confidence interval of [7.1, 9.9]. FactReview also reaches the highest accuracy at 79.2\%, 6.4 points above Codex.
\begin{figure}[t]
\centering
\includegraphics[width=\columnwidth]{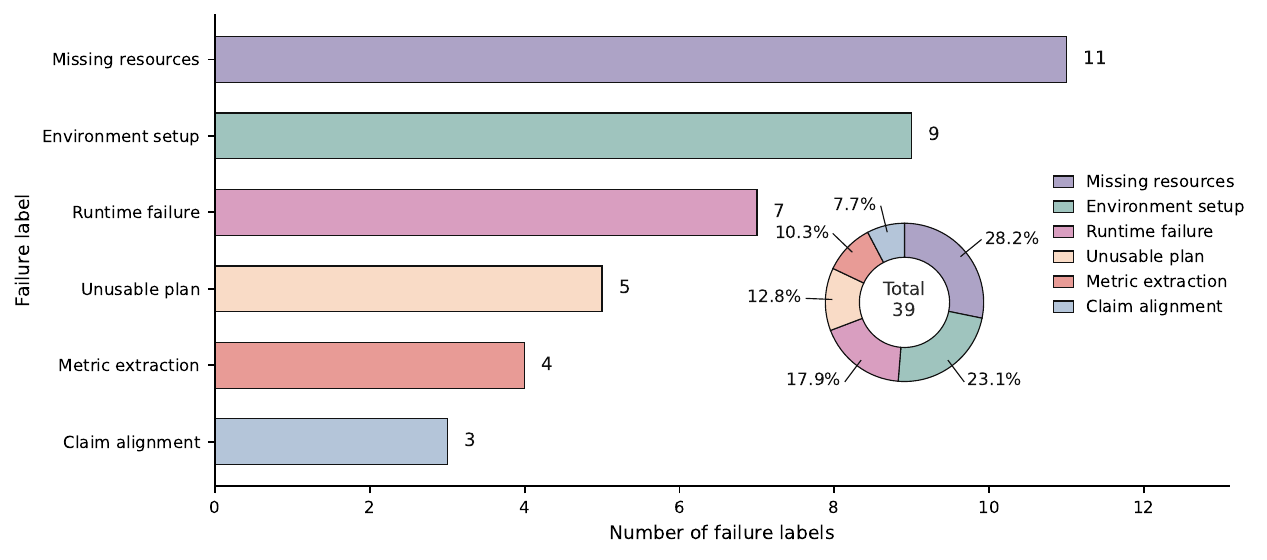}
\caption{Execution obstacles across the 28 papers with code. A paper can carry several labels, so the 39 labels exceed the paper count. Failures resolved through repair remain included in the counts.}
\label{fig:execution_failures}
\end{figure}

\noindent\textbf{Gate Levels.}
The gate sets how much evidence a claim needs to be labeled Supported, and a stricter gate leaves more claims with insufficient evidence Unverified. As shown in Figure~\ref{fig:gate_levels}, Lenient raises FactReview's coverage to 84.7\%, above Codex, while its false support rate stays lower at 10.4\%. Coverage and false support both decrease as the gate tightens. The main table reports Moderate results, and the levels are defined in Appendix~\ref{app:decision_rules}.

\noindent\textbf{Claim Extraction and Issue Detection.}
FactReview and the general purpose agents extract similar numbers of claims with similar F1, as shown in Figure~\ref{fig:claim_extraction}. After verification, FactReview's issue recall exceeds Codex by 5.8 percentage points, with fewer reference issues labeled Supported. Against concerns raised in public reviews, its recall exceeds Codex by 6.3 points, which agrees with the claim level comparison.

\noindent\textbf{Results by Paper Group.}
FactReview has lower false support than both general purpose agents in every group, as shown in Table~\ref{tab:group_results}. On AI generated papers, its false support rate is 10.5 percentage points below Codex. The recent group examines performance on research that models are less likely to have seen during training, which matches the setting of reviewing new submissions. In this group, FactReview's coverage is 78.2\%, compared with 81.4\% for Codex, and its false support rate is 8.0 points lower, close to the overall gap.

\subsection{Ablation Study}

FactReview and the general purpose agents extract claims similarly, yet their judgments differ after verification. The ablations examine how each verification branch, the execution module, and the status decision rules contribute to this gap.

Figure~\ref{fig:ablation_branches} shows that removing any verification branch lowers issue recall. Literature verification has the largest effect, with a decrease of 14.6 percentage points. Removing the experimental design check raises false support from 6.8\% to 12.7\% while coverage rises to 80.2\%. Code inspection applies only to papers with repositories, so its removal changes the aggregate results less.

\noindent\textbf{Execution Module.}
Figure~\ref{fig:ablation_execution} compares execution settings on the 28 papers with code, including those without successful execution. Removing execution verification lowers coverage from 80.0\% to 76.8\% and raises false support from 6.5\% to 8.0\%. Without the execution plan, coverage falls to 78.2\% and false support rises to 7.6\%, since the plan directs the execution agent toward experiments that test the claims under review. Disabling repair lowers both coverage and issue recall. On 56 planned tasks, valid tasks increase from 24 without repair to 36 after 3 rounds, a gain of 21.4 percentage points. The first round recovers the most tasks, and later rounds add progressively fewer, as shown in Figure~\ref{fig:repair_rounds}.

\noindent\textbf{Status Decision.}
This comparison holds the claims, the evidence, and each branch's sufficiency assessments fixed. The LLM receives the complete decision rules, and the same gate is applied to its output. Replacing programmatic assignment with the LLM still raises false support from 6.8\% to 12.4\% and lowers issue recall from 76.4\% to 71.3\%, while coverage rises to 80.9\%. Once evidence is collected, how the rules are applied still affects the reliability of the judgment. Table~\ref{tab:status_ablation} reports this comparison, and checklist sensitivity is examined in Appendix~\ref{app:checklist_sensitivity}.

\subsection{Case Study}

Among the 28 papers with code repositories, missing resources, environment failures, and runtime failures affect 11, 9, and 7 papers, respectively, with overlapping categories (Figure~\ref{fig:execution_failures}). After repair, 14 papers complete at least 1 experiment, and 8 provide valid execution evidence for at least 1 claim (Table~\ref{tab:execution_funnel}).

The case concerns a paper on text controlled generation of 3D urban scenes, from which FactReview extracts 116 verifiable claims. The paper attributes the largest contribution to generation quality to Urban MLLM, yet its ablation shows that removing this module increases FID by 4.49 while removing texture enhancement increases FID by 15.38, so the conclusion needs a narrower scope. Code inspection finds that rendering prompts are selected from predefined descriptions after receiving the user instruction. FactReview plans 7 experiments, including text controllability, but only 1 initially completes. Repairs fix incompatible dependencies and the launch directory, allowing the texture program to load the exported scene mesh. After 2 rounds, 3 experiments complete, including the text control check and FID evaluation.

The text control check runs the same scene objects with the instructions ``university campus'' and ``industrial district'' under a fixed seed. Both runs output textures, but the descriptions passed to the renderer are identical. Direct labels the text control claim Supported, while Codex and FactReview label it Questioned. In the FID evaluation, 2 runs give 392.46 and 390.18 against the reported 368.72, and checking the reference images and evaluation configuration does not explain the gap. FactReview labels the performance claim Questioned and retains the outputs for inspection. Of the paper's 36 reference issues, Direct supports 11, Codex 6, and FactReview 4.
\subsection{Human Evaluation}
\label{sec:human_evaluation}

The human evaluation draws a stratified sample of 36 papers from the 60 complete inventories. Participants are 10 master's or PhD students with reviewing experience. Each paper is assessed by 3 participants, and each participant receives 9 to 12 papers matched to their research area. Participants extract claims and assign statuses before seeing any output produced by the systems.

Human and FactReview claims are pooled, deduplicated, and adjudicated with their sources hidden. Against this inventory, average human claim recall is 68.3\%, compared with 88.1\% for FactReview, and participants accept 72.6\% of the additional claims FactReview extracts. On claims extracted by both, status agreement is 61.5\%. Participants label 53.6\% of their claims Unverified.
In a blinded comparison with evidence in a common format, FactReview covers 77.2\% of the confirmed issues in human reports, compared with 68.4\% for Codex. For findings absent from human reports, participants accept 84.6\% of FactReview's judgments after checking the manuscript and evidence, compared with 72.9\% for Codex.

The assisted review study compares independent review, Codex assistance, and FactReview assistance. Each paper receives 1 review per condition from different participants, who review papers they did not see earlier, and each condition contains 36 reviews. Reports are generated in advance, and timing runs from opening the paper to submitting the review, including report reading and evidence checking. With FactReview assistance, participants identify 9.4 reference issues per paper in 24.8 minutes, compared with 5.6 issues in 95.6 minutes and 7.1 issues in 33.7 minutes with Codex.

\noindent\textbf{Cost.}
Model calls cost \$3.2 per paper on average. For papers with a completed experiment, mean total cost including execution is \$14.4. Reviewers can choose which experiments to run and set the compute budget and repair limit. The breakdown is provided in Appendix~\ref{app:cost}.
\section{Conclusion}

This paper introduces FactReview, a framework for trustworthy AI peer review through literature checks, derivation analysis, and execution of available code. Across 300 papers, FactReview achieves leading overall performance among all evaluated systems, with the lowest false support rate of 6.8\%, the highest claim classification accuracy of 79.2\%, and the highest issue recall. In a study involving 36 papers, reviewers using FactReview identify more issues while spending 74.1\% less time than in independent review. These results demonstrate the value of evidence-grounded claim verification for faster, more trustworthy peer review.
\clearpage
\section*{Limitations}

Our evaluation covers multiple paper sources and two model backends, while broader disciplinary coverage remains an opportunity for future work. Execution results reflect the evaluated environment and resource budget, and may vary under different computational settings. The reviewer-assistance study involves participants with prior reviewing experience, so its findings may not fully generalize to reviewers with different levels of expertise.

\section*{Ethical Statement}

FactReview presents evidence-linked findings to reviewers and returns Inconclusive when evidence is insufficient. It uses public papers, repositories, and comments; participants are reported only in aggregate. Execution repairs dependencies, environments, paths, launch arguments, or wrappers, never scientific logic. Generative AI assisted with language and formatting; the authors retain responsibility.

\bibliography{arxiv_v5}
\clearpage
\raggedbottom

\appendix

\section{Implementation Details}
\label{app:implementation}

\subsection{Models, Budgets, and Environment}

\noindent\textbf{Models.} FactReview uses gpt-5.5. Claude Code uses Claude Opus 5.0, and the other baselines use gpt-5.5. Each system receives the same manuscript and available supplementary materials. Repository access is provided when code is available. Public review comments are withheld from all systems.

\noindent\textbf{Budgets.} Each system has a budget of 60 minutes and 10 million tokens per paper. Token accounting includes input, cached input, and output tokens across all stages. The time limit includes code execution. When either limit is reached, the system retains completed findings and leaves unresolved claims Unverified.

\noindent\textbf{Environment.} Execution uses an isolated Linux environment with 16 CPU cores, 64 GB RAM, and access to 1 NVIDIA A100 GPU with 80 GB memory. GPU resources are allocated when required by the experiment. Each paper has a separate workspace and environment.

\subsection{Execution and Repair}

\noindent\textbf{Commands and repairs.} Commands are derived from the repository documentation. Repairs may correct dependency versions or configuration paths. They may also adjust the working directory or launch arguments. The scientific implementation remains unchanged. Each repair is recorded alongside the failed command and the resulting file difference in the execution record.

At most 3 repair rounds are allowed within the shared budget. Each round diagnoses a failed attempt, applies an allowed repair, and evaluates the revised command. Downloading a required public resource counts toward the same budget.

\noindent\textbf{Execution evidence.} An execution task specifies a target claim and an experiment that can assess it. The plan records the expected output and the conditions required for comparison with the paper. These conditions are checked before interpreting a numerical difference.

A result becomes usable evidence when the required output is recovered and the execution conditions match the claim. The execution judge returns pass when the result supports the claim, fail when aligned evidence contradicts it, and inconclusive when the output does not resolve it. These outcomes inform the final claim decision.

\noindent\textbf{Variation.} When the paper reports uncertainty, the comparison uses that uncertainty. Otherwise, the plan specifies the repeated runs or additional checks needed to assess variation. A numerical discrepancy without adequate information about variation may justify Questioned.

\subsection{Status Assignment and Gate}
\label{app:decision_rules}

\noindent\textbf{Statuses.} Supported requires evidence covering the stated claim. Flawed requires a demonstrated contradiction or error. Questioned requires a specific concern supported by evidence, with part of the claim unresolved. Unverified applies when the available evidence is insufficient to reach any of the other three decisions.

\noindent\textbf{Gate.} The gate determines whether the supporting evidence is sufficient for Supported. It uses the sufficiency assessments returned by the checks assigned to a claim. Inapplicable checks are excluded from the counts below.

Let $m$ be the number of applicable checks and $n$ the number returning sufficient supporting evidence. Lenient requires $n\geq 1$. Moderate requires $n\geq\lceil 2m/3\rceil$. Strict requires $n=m$. A claim with $m=0$ remains Unverified.

All levels retain the same checks for unresolved contradictions and scope mismatch. Passing the gate permits Supported only after these checks pass. The applicable checks for each claim are fixed before comparing the three gate levels.

\subsection{Baseline Outputs}

\noindent\textbf{Report conversion.} Direct extracts claims and assigns statuses in a single model call. The dedicated reviewing systems produce reports in their native formats. A common conversion step identifies the claims addressed by each report and retains the evidence stated there.

The conversion step does not retrieve additional evidence. A demonstrated error maps to Flawed. A supported concern that remains unresolved maps to Questioned. A favorable assessment maps to Supported when it addresses the corresponding claim. Claims absent from a report remain Unverified.

\noindent\textbf{Agent baselines.} Claude Code and Codex receive the same verification instructions as FactReview. Their tool use follows their native workflows. Appendix~\ref{app:backend_results} compares FactReview and Direct under the same backend.

\section{Dataset and Evaluation Protocol}
\label{app:dataset}

\subsection{Dataset Composition}

\noindent\textbf{Groups.} The dataset contains 300 papers in 6 mutually exclusive groups. Papers assigned to the recent group are excluded from the other groups. Different versions of the same paper count as a single paper. Table~\ref{tab:dataset_denominators} gives the group sizes and evaluation denominators.

\begin{table*}[t]
\centering\small
\setlength{\tabcolsep}{5pt}
\begin{tabular}{llrrrrr}
\toprule
Group & Source & Papers & Complete & Light & $N_S$ & $N_P$ \\
\midrule
AI generated & Agents4Science 2025 & 52 & 12 & 40 & 615 & 776 \\
ICLR accepted & ICLR 2025 & 52 & 12 & 40 & 988 & 430 \\
ICLR rejected & ICLR 2025 & 52 & 12 & 40 & 704 & 603 \\
TMLR & TMLR & 52 & 12 & 40 & 810 & 523 \\
Nature Communications & Nature Communications & 26 & 6 & 20 & 356 & 296 \\
Recent papers & Recent submissions & 66 & 6 & 60 & 560 & 422 \\
\midrule
Total & & 300 & 60 & 240 & 4033 & 3050 \\
\bottomrule
\end{tabular}
\caption{Dataset composition. Complete denotes papers with a complete claim inventory. Light denotes papers annotated for selected claims. $N_S$ counts reference Supported claims. $N_P$ counts reference Flawed or Questioned claims, which serve as the denominator of false support and issue recall.}
\label{tab:dataset_denominators}
\end{table*}

\noindent\textbf{Sampling.} Sampling is stratified by source. Reviewer issue density informs the sampling weights within each source. Code availability is also considered when selecting papers for execution analysis. Every system is evaluated on the same selected papers and reference claims.

\noindent\textbf{Recent papers.} The recent group contains papers whose first public version appeared after January 31, 2026. The date is taken from the earliest available arXiv or OpenReview version. Later revisions do not change group membership. This group examines performance on more recent research. Publication dates alone do not establish whether a paper was excluded from model training.

\subsection{Annotation and Public Review Issues}
\label{app:annotation}

\noindent\textbf{Claim annotation.} FactReview's preparation pipeline proposes candidate claims. Annotators check them against the full manuscript. They remove invalid entries, merge duplicates, and add omitted claims. They assign statuses using the evidence reviewed during annotation.

The 60 complete inventories contain 7300 reference claims. Of these, 1900 are Unverified, and 5400 are assigned another status. The 240 lightly annotated papers contain 2103 reference claims, including 420 Unverified claims. The remaining 1683 claims contribute to the Supported and issue denominators of Table~\ref{tab:dataset_denominators}.

Light annotation covers selected claims affecting the main contribution or linked to a specific concern. These papers contribute to status evaluation. Claim extraction recall uses only the 60 complete inventories.

A core claim directly supports the paper's main contribution or principal findings. Core status is assigned before inspecting system predictions. Evidence sufficiency is annotated separately from claim status. Each annotation retains the source location supporting the decision.

\noindent\textbf{Agreement.} The agreement sample is drawn across paper groups. Its claims are labeled independently by 2 annotators. Table~\ref{tab:annotation_agreement} reports their agreement before adjudication. The human study separately compares participants' judgments with FactReview outputs in Section~\ref{sec:human_evaluation}.

\begin{table}[t]
\centering\small
\setlength{\tabcolsep}{5pt}
\begin{tabular}{lr}
\toprule
Measure & Value \\
\midrule
Independently double annotated claims & 500 \\
Status agreement (\%) & 82.6 \\
Cohen's $\kappa$ & 0.757 \\
\bottomrule
\end{tabular}
\caption{Annotation agreement before adjudication. A third annotator resolves disagreements.}
\label{tab:annotation_agreement}
\end{table}

\noindent\textbf{Public review issues.}\label{app:public_reviews} The public review reference contains specific concerns raised by human reviewers. General recommendations and comments without an identifiable concern are excluded. Repeated concerns are merged when they address the same claim and require the same evidence.

Each retained issue is linked to a manuscript passage. The reference preserves the reviewer's concern without treating it as an adjudicated error. Reviewer issue recall measures whether a system identifies the same concern.

The eligible set contains 222 papers. Table~\ref{tab:public_review_subset} gives its composition. Papers without an eligible public review issue are excluded from this metric.

\begin{table}[t]
\centering\small
\setlength{\tabcolsep}{5pt}
\begin{tabular}{lr}
\toprule
Group & Eligible papers \\
\midrule
AI generated & 52 \\
ICLR accepted & 52 \\
ICLR rejected & 52 \\
TMLR & 34 \\
Nature Communications & 0 \\
Recent papers & 32 \\
\midrule
Total & 222 \\
\bottomrule
\end{tabular}
\caption{Papers contributing to reviewer issue recall. Each eligible paper has at least 1 specific issue from a public human review.}
\label{tab:public_review_subset}
\end{table}

\subsection{Claim Matching}
\label{app:matching}

\noindent\textbf{Candidate pairs.} Claims are matched within each paper. Semantic similarity retrieves candidate pairs. A separate judgment checks whether both statements express the same proposition under the same conditions. System names and predicted statuses are hidden during matching.

\noindent\textbf{Scope.} A match requires agreement on claim scope. A result under narrower conditions does not match a broader claim when the omitted conditions could change its truth. Compound statements are split before matching, and duplicates are merged.

\noindent\textbf{Assignment.} Accepted pairs form a bipartite graph between system and reference claims. Maximum cardinality matching assigns each claim to at most 1 counterpart. Ties are resolved by total semantic similarity. Unmatched reference claims receive an Unverified prediction for status evaluation.

\noindent\textbf{Audit.} An audit of 600 candidate pairs gives matching precision of 96.4\% and recall of 94.1\%. The audit includes rejected candidates to measure missed matches. Matching settings remain fixed across all systems and paper groups.

\subsection{Metric Definitions}
\label{app:metrics}

\noindent\textbf{Status metrics.} Let $S$, $F$, $Q$, and $U$ denote Supported, Flawed, Questioned, and Unverified. For reference claim $i$, let $y_i$ be its adjudicated status and $\hat y_i$ its matched system status. Define
\[
\mathcal{S}=\{i:y_i=S\},\qquad
\mathcal{P}=\{i:y_i\in\{F,Q\}\}.
\]
False support rate, coverage, and issue recall are
\[
\begin{aligned}
\mathrm{FSR}
&=\frac{\sum_{i\in\mathcal{P}}\mathbf{1}[\hat y_i=S]}
{|\mathcal{P}|},\\
\mathrm{Cov}
&=\frac{\sum_{i\in\mathcal{S}}\mathbf{1}[\hat y_i=S]}
{|\mathcal{S}|},\\
\mathrm{IR}
&=\frac{\sum_{i\in\mathcal{P}}\mathbf{1}[\hat y_i\in\{F,Q\}]}
{|\mathcal{P}|}.
\end{aligned}
\]
Core issue recall restricts this calculation to core claims. Four status accuracy requires an exact match between $y_i$ and $\hat y_i$. Predicting Questioned for a reference Flawed claim contributes to issue recall but is an incorrect status prediction.

\noindent\textbf{Extraction metrics.} Let $M$ be the number of matched claims. Let $N_{\mathrm{pred}}$ and $N_{\mathrm{ref}}$ be the extracted and reference claim counts. Then
\[
P=\frac{M}{N_{\mathrm{pred}}},\qquad
R=\frac{M}{N_{\mathrm{ref}}},\qquad
F_1=\frac{2PR}{P+R}.
\]
Counts are summed over the 60 complete inventories before computing the extraction scores.

\noindent\textbf{Reviewer issue recall.} Reviewer issue recall divides matched public review issues by all eligible reference issues. Matching requires the same underlying concern. Mentioning the same topic without the same concern is insufficient.

\noindent\textbf{Aggregation.} Aggregate rates are computed from summed counts. Group percentages are rounded after computation. Gate comparisons retain the same reference claims at every level. The false support axis measures errors among reference issues, while the coverage axis measures recovered reference Supported claims.

\section{Additional Results}
\label{app:additional_experiments}

\subsection{Results by Paper Group}

Table~\ref{tab:group_results} reports the results for each paper group. Table~\ref{tab:aggregation_counts} reports the numerators underlying the group comparison. Correct support uses the Supported denominators in Table~\ref{tab:dataset_denominators}. False support and detected issues use the issue denominators.

\begin{table*}[t]
\centering\small
\setlength{\tabcolsep}{7pt}
\renewcommand{\arraystretch}{1.12}
\begin{tabular}{llccc}
\toprule
Paper group & System & FSR $\downarrow$ & Cov. $\uparrow$ & Issue R. $\uparrow$ \\
\midrule
AI generated & Claude Code & 21.1 & \underline{74.1} & \underline{73.2} \\
& Codex & \underline{18.9} & \textbf{75.9} & 72.8 \\
\rowcolor{gray!12}
& \textbf{FactReview} & \textbf{8.4} & 72.0 & \textbf{79.9} \\
\midrule
ICLR accepted & Claude Code & \underline{10.9} & \underline{80.7} & 60.7 \\
& Codex & 12.1 & \textbf{82.8} & \underline{64.4} \\
\rowcolor{gray!12}
& \textbf{FactReview} & \textbf{4.4} & 78.5 & \textbf{69.3} \\
\midrule
ICLR rejected & Claude Code & 17.6 & \underline{82.2} & 75.8 \\
& Codex & \underline{15.1} & \textbf{84.1} & \underline{77.8} \\
\rowcolor{gray!12}
& \textbf{FactReview} & \textbf{6.6} & 79.0 & \textbf{83.6} \\
\midrule
TMLR & Claude Code & 14.1 & \textbf{83.8} & 67.5 \\
& Codex & \underline{11.9} & \underline{83.7} & \underline{71.3} \\
\rowcolor{gray!12}
& \textbf{FactReview} & \textbf{5.5} & 80.6 & \textbf{76.7} \\
\midrule
Nature Communications & Claude Code & \underline{18.2} & \textbf{81.2} & \underline{62.5} \\
& Codex & 18.9 & \underline{79.8} & 61.5 \\
\rowcolor{gray!12}
& \textbf{FactReview} & \textbf{9.5} & 74.2 & \textbf{66.9} \\
\midrule
Recent papers & Claude Code & 15.6 & \underline{79.1} & 65.6 \\
& Codex & \underline{14.2} & \textbf{81.4} & \underline{68.0} \\
\rowcolor{gray!12}
& \textbf{FactReview} & \textbf{6.2} & 78.2 & \textbf{73.2} \\
\bottomrule
\end{tabular}
\caption{Results by paper group. All values are percentages. Each group uses the same reference claims across systems. Aggregate scores use the counts in Tables~\ref{tab:dataset_denominators} and~\ref{tab:aggregation_counts}. Best results are in bold. Second-best results are underlined, and comparisons are made within each paper group.}
\label{tab:group_results}
\end{table*}

\noindent\textbf{Aggregate scores.} FactReview's aggregate coverage and false support rate are
\[
\begin{aligned}
\mathrm{Cov} &= \frac{3130}{4033}=77.6\%,\\
\mathrm{FSR} &= \frac{207}{3050}=6.8\%.
\end{aligned}
\]
The same aggregation gives Codex coverage of 81.7\% and false support of 15.3\%.

\begin{table*}[t]
\centering\small
\setlength{\tabcolsep}{7pt}
\begin{tabular}{llrrr}
\toprule
Group & System & Correct support & False support & Detected issues \\
\midrule
AI generated & Claude Code & 456 & 164 & 568 \\
& Codex & 467 & 147 & 565 \\
\rowcolor{gray!12}
& FactReview & 443 & 65 & 620 \\
\midrule
ICLR accepted & Claude Code & 797 & 47 & 261 \\
& Codex & 818 & 52 & 277 \\
\rowcolor{gray!12}
& FactReview & 776 & 19 & 298 \\
\midrule
ICLR rejected & Claude Code & 579 & 106 & 457 \\
& Codex & 592 & 91 & 469 \\
\rowcolor{gray!12}
& FactReview & 556 & 40 & 504 \\
\midrule
TMLR & Claude Code & 679 & 74 & 353 \\
& Codex & 678 & 62 & 373 \\
\rowcolor{gray!12}
& FactReview & 653 & 29 & 401 \\
\midrule
Nature Communications & Claude Code & 289 & 54 & 185 \\
& Codex & 284 & 56 & 182 \\
\rowcolor{gray!12}
& FactReview & 264 & 28 & 198 \\
\midrule
Recent papers & Claude Code & 443 & 66 & 277 \\
& Codex & 456 & 60 & 287 \\
\rowcolor{gray!12}
& FactReview & 438 & 26 & 309 \\
\midrule
Total & Claude Code & 3243 & 511 & 2101 \\
& Codex & 3295 & 468 & 2153 \\
\rowcolor{gray!12}
& FactReview & 3130 & 207 & 2330 \\
\bottomrule
\end{tabular}
\caption{Counts underlying the group results. Every system uses the same reference denominators within each paper group, which are listed in Table~\ref{tab:dataset_denominators}.}
\label{tab:aggregation_counts}
\end{table*}

\subsection{Backend and Research Area}
\label{app:backend_results}

\noindent\textbf{Backend comparison.} The comparison uses the same 60 papers with complete claim inventories. FactReview and Direct use each backend under the same budget. Reference annotations and claim matching remain fixed. Across both backends, FactReview reduces false support and improves issue recall while maintaining coverage close to that of Direct, as shown in Table~\ref{tab:backend_results}.

\begin{table*}[t]
\centering\small
\setlength{\tabcolsep}{8pt}
\begin{tabular}{llccc}
\toprule
Backend & System & FSR $\downarrow$ & Cov. $\uparrow$ & Issue R. $\uparrow$ \\
\midrule
gpt-5.5 & Direct & 30.2 & 77.3 & 45.1 \\
\rowcolor{gray!12}
& FactReview & 7.2 & 78.1 & 77.0 \\
\midrule
Claude Opus 5.0 & Direct & 26.8 & 75.9 & 47.2 \\
\rowcolor{gray!12}
& FactReview & 8.5 & 76.4 & 75.3 \\
\bottomrule
\end{tabular}
\caption{Backend comparison on the 60 papers with complete claim inventories. All values are percentages.}
\label{tab:backend_results}
\end{table*}

\noindent\textbf{Research area.}\label{app:subfield_results} The 60 complete inventories are grouped by each paper's primary research area. General learning methods include contributions that are not specific to language or vision. Each paper appears in 1 group. FactReview has lower false support and higher issue recall in each area (Table~\ref{tab:subfield_results}). Vision papers have the lowest coverage for both systems.

\begin{table*}[t]
\centering\small
\setlength{\tabcolsep}{7pt}
\begin{tabular}{lrlccc}
\toprule
Area & Papers & System & FSR $\downarrow$ & Cov. $\uparrow$ & Issue R. $\uparrow$ \\
\midrule
Language & 18 & Codex & 14.1 & 83.5 & 73.2 \\
\rowcolor{gray!12}
& & FactReview & 6.3 & 79.8 & 78.5 \\
\midrule
Vision & 18 & Codex & 17.2 & 80.1 & 68.4 \\
\rowcolor{gray!12}
& & FactReview & 8.2 & 76.3 & 75.3 \\
\midrule
General learning methods & 24 & Codex & 15.5 & 82.0 & 72.1 \\
\rowcolor{gray!12}
& & FactReview & 7.1 & 78.2 & 77.1 \\
\bottomrule
\end{tabular}
\caption{Results by research area on the 60 papers with complete inventories. All metric values are percentages.}
\label{tab:subfield_results}
\end{table*}

\subsection{Status Assignment and Checklist Sensitivity}

\noindent\textbf{Status assignment.} The status comparison fixes the claims and collected evidence. Evidence sufficiency assessments and the gate also remain unchanged. The LLM receives the complete decision rules. The program applies these rules to the same records. Table~\ref{tab:status_ablation} shows that the LLM achieves higher coverage alongside more false support and lower issue recall. This comparison isolates status assignment from evidence collection.

\begin{table}[t]
\centering\small
\setlength{\tabcolsep}{3pt}
\begin{tabular}{lccc}
\toprule
Decision method & FSR $\downarrow$ & Cov. $\uparrow$ & Issue R. $\uparrow$ \\
\midrule
LLM with complete rules & 12.4 & 80.9 & 71.3 \\
\rowcolor{gray!12}
FactReview & 6.8 & 77.6 & 76.4 \\
\bottomrule
\end{tabular}
\caption{Status assignment on the same claims and evidence from all 300 papers. All values are percentages.}
\label{tab:status_ablation}
\end{table}

\noindent\textbf{Checklist sensitivity.}\label{app:checklist_sensitivity} Checklist sensitivity uses the same 60 complete inventories. Reference claims and status rules remain fixed. A shared checklist applies the same questions to every paper. The standard configuration selects checks from the paper's claims. An expanded checklist adds checks for secondary claims. All settings use the same budget of time and tokens.

The shared checklist misses more reference issues (Table~\ref{tab:checklist_sensitivity}). Expanding the standard checklist increases issue recall by 1.3 percentage points, while coverage decreases by 1.5 points.

\begin{table}[t]
\centering\small
\setlength{\tabcolsep}{4pt}
\begin{tabular}{lccc}
\toprule
Checklist & FSR $\downarrow$ & Cov. $\uparrow$ & Issue R. $\uparrow$ \\
\midrule
Shared & 9.1 & \underline{77.0} & 70.4 \\
\rowcolor{gray!12}
Standard & \textbf{7.2} & \textbf{78.1} & \underline{77.0} \\
Expanded & \underline{7.8} & 76.6 & \textbf{78.3} \\
\bottomrule
\end{tabular}
\caption{Checklist sensitivity with reference claims and decision rules held fixed. All values are percentages. Best results are in bold. Second-best results are underlined in each column.}
\label{tab:checklist_sensitivity}
\end{table}

\subsection{Execution, Repair, and Cost}
\label{app:execution_details}

\noindent\textbf{Execution subset.} The execution analysis includes all 28 papers with code repositories. Papers remain in the evaluation when their environments cannot be prepared or their experiments fail. Of these papers, 12 have complete claim inventories. The subset contains 680 reference Supported claims and 510 reference issues.

\noindent\textbf{Failure categories.} Failure categories record obstacles encountered at any attempt. A resolved obstacle remains in the category count. Repeated occurrences of the same category within a paper count once toward that category.

An unusable plan fails to specify an executable experiment for the target claim. Missing resources prevent the intended experiment from starting. Environment failures occur during setup, while runtime failures interrupt an initiated experiment. Metric extraction fails when the expected result cannot be recovered. An alignment mismatch occurs when the recovered result uses conditions different from those required by the claim.

\noindent\textbf{Execution progress.} Table~\ref{tab:execution_funnel} records final progress after repair. The stages are cumulative. A repaired obstacle can therefore appear in the failure analysis without blocking the paper from reaching the final execution stage.

\begin{table}[t]
\centering\small
\setlength{\tabcolsep}{4pt}
\begin{tabular}{p{0.76\columnwidth}r}
\toprule
Execution stage & Papers \\
\midrule
Code repository available & 28 \\
Required resources available and environment configured & 19 \\
At least 1 experiment completed & 14 \\
Target metric obtained & 11 \\
Result aligned with the claim and usable as evidence & 8 \\
\bottomrule
\end{tabular}
\caption{Execution progress after repair. Each stage includes only papers that passed every earlier stage.}
\label{tab:execution_funnel}
\end{table}

\noindent\textbf{Repair.}\label{app:repair_analysis} Repair budgets are compared on the same 56 planned tasks. Repeated attempts do not create new tasks. As reported in Figure~\ref{fig:repair_rounds}, usable evidence increases from 24 tasks without repair to 36 after 3 rounds. Successive rounds recover 7, 4, and 1 additional tasks. The final valid tasks are distributed across 8 papers. Task success and paper success therefore have different denominators.

The final paper funnel leaves 9 papers blocked during preparation, 5 during execution, 3 during metric recovery, and 3 during alignment. The remaining 8 papers provide usable evidence. These categories are exclusive and sum to 28.

\noindent\textbf{Cost.}\label{app:cost} Model cost is calculated from recorded token use and the rates applied to each run. Cached input is charged at its applicable rate. Execution cost includes compute used by failed attempts and repairs. Table~\ref{tab:cost_breakdown} separates average model cost across all 300 papers from average total cost for the 14 papers completing at least 1 experiment. These averages use different subsets.

\begin{table}[t]
\centering\small
\setlength{\tabcolsep}{4pt}
\begin{tabular}{p{0.67\columnwidth}r}
\toprule
Cost component & USD per paper \\
\midrule
\multicolumn{2}{l}{All 300 papers} \\
Claim extraction & 0.28 \\
Literature verification & 1.17 \\
Theory and design checks & 0.83 \\
Code analysis and execution coordination & 0.54 \\
Report generation & 0.38 \\
Total model cost & 3.20 \\
\midrule
\multicolumn{2}{l}{Papers with a completed experiment} \\
Model calls & 4.10 \\
Execution compute & 10.30 \\
Total cost & 14.40 \\
\bottomrule
\end{tabular}
\caption{Average FactReview cost. The upper block excludes execution compute. The lower block includes the total cost for papers with a completed experiment.}
\label{tab:cost_breakdown}
\end{table}

Literature verification accounts for the largest share of model cost. For papers with completed experiments, execution compute accounts for 71.5\% of total cost. Reviewers can select the experiments to run and set a compute budget. The repair limit controls the number of additional execution attempts allowed for each plan.

\section{Human Evaluation}
\label{app:human_evaluation}

\subsection{Participants and Papers}

\noindent\textbf{Papers.} The human study uses 36 papers drawn from the 60 complete inventories. The allocation contains 8 papers from each ICLR group and 8 AI generated papers. It also contains 6 TMLR papers, with 3 papers from each remaining group.

\noindent\textbf{Participants.} The participants are 10 master's or PhD students with prior reviewing experience. Assignments account for research familiarity. Participants do not assess papers with which they have a known conflict of interest.

\noindent\textbf{Instructions.} Participants assess the paper's claims and explain each reported concern. In assisted conditions, the system report is available alongside the manuscript. Participants inspect the evidence for findings included in their reviews. During report evaluation, a finding is accepted when its evidence supports the stated concern or conclusion. Unsupported findings are rejected. Findings that cannot be resolved from the supplied material remain uncertain and count as not accepted.

\subsection{Claim Assessment}

\noindent\textbf{Procedure.} Each paper is assessed independently by 3 participants. Each participant receives 9 to 12 papers in this stage. Participants extract verifiable claims before seeing system outputs and assign statuses using the available evidence.

\noindent\textbf{Pooled inventory.} Human and FactReview claims are pooled and deduplicated. Adjudicators review the pooled records with their extraction sources hidden. Invalid claims are removed. The resulting inventory is used for this human comparison.

\noindent\textbf{Measures.} Human recall is computed for each participant before averaging. FactReview recall uses the same pooled inventory. Status agreement is calculated on matched claims. The conditional Unverified rate includes only claims labeled Unverified by the participant. Table~\ref{tab:human_claim_assessment} reports the results of this comparison.

\begin{table}[t]
\centering\small
\setlength{\tabcolsep}{4pt}
\begin{tabular}{p{0.76\columnwidth}r}
\toprule
Measure & Value (\%) \\
\midrule
Human claim recall & 68.3 \\
FactReview claim recall & 88.1 \\
Acceptance of additional extracted claims & 72.6 \\
Human and FactReview status agreement & 61.5 \\
Human Unverified rate & 53.6 \\
FactReview Unverified among human Unverified claims & 31.2 \\
\bottomrule
\end{tabular}
\caption{Human claim assessment. Additional extracted claims are FactReview claims absent from the evaluating participant's extraction.}
\label{tab:human_claim_assessment}
\end{table}

\subsection{Assisted Review and Report Evaluation}

\noindent\textbf{Assignment.} Each paper receives 1 review in each condition, completed by different participants. Participants review papers they did not encounter during claim assessment. Each participant completes 3 or 4 reviews per condition. Condition order is balanced across participants. Source composition is identical across conditions because every paper appears in every condition. Outcome comparisons are paired by paper.

\noindent\textbf{Timing and findings.} Reports are generated before the session. Timing begins when the participant opens the paper and ends when the review is submitted. It includes report reading and evidence checking. Recorded breaks are excluded. Participants provide a manuscript location or external evidence for each finding. Adjudicators match submitted findings to reference issues with the assistance condition hidden. Repeated descriptions of the same issue count once.

\noindent\textbf{Report evaluation.} Human reports are the reviews written in the independent condition. Their findings are adjudicated against the cited evidence. A confirmed Questioned finding establishes that the concern is supported. FactReview and Codex findings use a common presentation format with system names removed. Findings are presented in randomized order. Participants evaluate papers they have not assessed in earlier stages.

Each finding is first judged against the paper. Its supplied evidence is then revealed for a second judgment. Participants assess evidence validity and whether it justifies the assigned status. Final acceptance requires both judgments to be positive.

\noindent\textbf{Coverage and acceptance.} Coverage uses confirmed human findings as its denominator. Additional findings are system findings without a match in the human report. Each additional finding receives 3 independent judgments. Acceptance is calculated over these judgments (Table~\ref{tab:report_acceptance}).

\begin{table}[t]
\centering\small
\setlength{\tabcolsep}{4pt}
\begin{tabular}{lcc}
\toprule
System & \shortstack{Human issue\\coverage (\%)} & \shortstack{Additional finding\\acceptance (\%)} \\
\midrule
Codex & 68.4 & 72.9 \\
\rowcolor{gray!12}
FactReview & 77.2 & 84.6 \\
\bottomrule
\end{tabular}
\caption{Coverage of confirmed human findings and acceptance of additional system findings after evidence inspection. Acceptance is calculated over individual participant judgments.}
\label{tab:report_acceptance}
\end{table}

\subsection{Statistical Analysis}
\label{app:statistical_analysis}

\noindent\textbf{Tests.} Human review outcomes are summarized by their mean and sample standard deviation. The primary contrasts compare FactReview assistance with independent review and Codex assistance. Confidence intervals use 20,000 bootstrap resamples of papers. Every sampled paper retains its outcomes from all conditions. Two sided tests use 50,000 random sign permutations of paired differences. Holm's method adjusts the p-values of the 4 comparisons for multiplicity.

\section{Dataset and Evaluation Protocol}
\label{app:dataset}

\subsection{Dataset Composition}

\noindent\textbf{Groups.} The dataset contains 300 papers in 6 mutually exclusive groups. Papers assigned to the recent group are excluded from the other groups. Different versions of the same paper count as a single paper. Table~\ref{tab:dataset_denominators} gives the group sizes and evaluation denominators.

\begin{table*}[t]
\centering\small
\setlength{\tabcolsep}{5pt}
\begin{tabular}{llrrrrr}
\toprule
Group & Source & Papers & Complete & Light & $N_S$ & $N_P$ \\
\midrule
AI generated & Agents4Science 2025 & 52 & 12 & 40 & 615 & 776 \\
ICLR accepted & ICLR 2025 & 52 & 12 & 40 & 988 & 430 \\
ICLR rejected & ICLR 2025 & 52 & 12 & 40 & 704 & 603 \\
TMLR & TMLR & 52 & 12 & 40 & 810 & 523 \\
Nature Communications & Nature Communications & 26 & 6 & 20 & 356 & 296 \\
Recent papers & Recent submissions & 66 & 6 & 60 & 560 & 422 \\
\midrule
Total & & 300 & 60 & 240 & 4033 & 3050 \\
\bottomrule
\end{tabular}
\caption{Dataset composition. Complete denotes papers with a complete claim inventory. Light denotes papers annotated for selected claims. $N_S$ counts reference Supported claims. $N_P$ counts reference Flawed or Questioned claims, which serve as the denominator of false support and issue recall.}
\label{tab:dataset_denominators}
\end{table*}

\noindent\textbf{Sampling.} Sampling is stratified by source. Reviewer issue density informs the sampling weights within each source. Code availability is also considered when selecting papers for execution analysis. Every system is evaluated on the same selected papers and reference claims.

\noindent\textbf{Recent papers.} The recent group contains papers whose first public version appeared after January 31, 2026. The date is taken from the earliest available arXiv or OpenReview version. Later revisions do not change group membership. This group examines performance on more recent research. Publication dates alone do not establish whether a paper was excluded from model training.

\subsection{Annotation and Public Review Issues}
\label{app:annotation}

\noindent\textbf{Claim annotation.} FactReview's preparation pipeline proposes candidate claims. Annotators check them against the full manuscript. They remove invalid entries, merge duplicates, and add omitted claims. They assign statuses using the evidence reviewed during annotation.

The 60 complete inventories contain 7300 reference claims. Of these, 1900 are Unverified, and 5400 are assigned another status. The 240 lightly annotated papers contain 2103 reference claims, including 420 Unverified claims. The remaining 1683 claims contribute to the Supported and issue denominators of Table~\ref{tab:dataset_denominators}.

Light annotation covers selected claims affecting the main contribution or linked to a specific concern. These papers contribute to status evaluation. Claim extraction recall uses only the 60 complete inventories.

A core claim directly supports the paper's main contribution or principal findings. Core status is assigned before inspecting system predictions. Evidence sufficiency is annotated separately from claim status. Each annotation retains the source location supporting the decision.

\noindent\textbf{Agreement.} The agreement sample is drawn across paper groups. Its claims are labeled independently by 2 annotators. Table~\ref{tab:annotation_agreement} reports their agreement before adjudication. The human study separately compares participants' judgments with FactReview outputs in Section~\ref{sec:human_evaluation}.

\begin{table}[t]
\centering\small
\setlength{\tabcolsep}{5pt}
\begin{tabular}{lr}
\toprule
Measure & Value \\
\midrule
Independently double annotated claims & 500 \\
Status agreement (\%) & 82.6 \\
Cohen's $\kappa$ & 0.757 \\
\bottomrule
\end{tabular}
\caption{Annotation agreement before adjudication. A third annotator resolves disagreements.}
\label{tab:annotation_agreement}
\end{table}

\noindent\textbf{Public review issues.}\label{app:public_reviews} The public review reference contains specific concerns raised by human reviewers. General recommendations and comments without an identifiable concern are excluded. Repeated concerns are merged when they address the same claim and require the same evidence.

Each retained issue is linked to a manuscript passage. The reference preserves the reviewer's concern without treating it as an adjudicated error. Reviewer issue recall measures whether a system identifies the same concern.

The eligible set contains 222 papers. Table~\ref{tab:public_review_subset} gives its composition. Papers without an eligible public review issue are excluded from this metric.

\begin{table}[t]
\centering\small
\setlength{\tabcolsep}{5pt}
\begin{tabular}{lr}
\toprule
Group & Eligible papers \\
\midrule
AI generated & 52 \\
ICLR accepted & 52 \\
ICLR rejected & 52 \\
TMLR & 34 \\
Nature Communications & 0 \\
Recent papers & 32 \\
\midrule
Total & 222 \\
\bottomrule
\end{tabular}
\caption{Papers contributing to reviewer issue recall. Each eligible paper has at least 1 specific issue from a public human review.}
\label{tab:public_review_subset}
\end{table}

\subsection{Claim Matching}
\label{app:matching}

\noindent\textbf{Candidate pairs.} Claims are matched within each paper. Semantic similarity retrieves candidate pairs. A separate judgment checks whether both statements express the same proposition under the same conditions. System names and predicted statuses are hidden during matching.

\noindent\textbf{Scope.} A match requires agreement on claim scope. A result under narrower conditions does not match a broader claim when the omitted conditions could change its truth. Compound statements are split before matching, and duplicates are merged.

\noindent\textbf{Assignment.} Accepted pairs form a bipartite graph between system and reference claims. Maximum cardinality matching assigns each claim to at most 1 counterpart. Ties are resolved by total semantic similarity. Unmatched reference claims receive an Unverified prediction for status evaluation.

\noindent\textbf{Audit.} An audit of 600 candidate pairs gives matching precision of 96.4\% and recall of 94.1\%. The audit includes rejected candidates to measure missed matches. Matching settings remain fixed across all systems and paper groups.

\subsection{Metric Definitions}
\label{app:metrics}

\noindent\textbf{Status metrics.} Let $S$, $F$, $Q$, and $U$ denote Supported, Flawed, Questioned, and Unverified. For reference claim $i$, let $y_i$ be its adjudicated status and $\hat y_i$ its matched system status. Define
\[
\begin{aligned}
\mathrm{FSR}
&= \frac{\sum_{i\in\mathcal{P}}\mathbf{1}[\hat y_i=S]}
         {|\mathcal{P}|}, \\[4pt]
\mathrm{Cov}
&= \frac{\sum_{i\in\mathcal{S}}\mathbf{1}[\hat y_i=S]}
         {|\mathcal{S}|}, \\[4pt]
\mathrm{IR}
&= \frac{\sum_{i\in\mathcal{P}}\mathbf{1}[\hat y_i\in\{F,Q\}]}
         {|\mathcal{P}|}.
\end{aligned}
\]
Core issue recall restricts this calculation to core claims. Four status accuracy requires an exact match between $y_i$ and $\hat y_i$. Predicting Questioned for a reference Flawed claim contributes to issue recall but is an incorrect status prediction.

\noindent\textbf{Extraction metrics.} Let $M$ be the number of matched claims. Let $N_{\mathrm{pred}}$ and $N_{\mathrm{ref}}$ be the extracted and reference claim counts. Then
\[
P=\frac{M}{N_{\mathrm{pred}}},\qquad
R=\frac{M}{N_{\mathrm{ref}}},\qquad
F_1=\frac{2PR}{P+R}.
\]
Counts are summed over the 60 complete inventories before computing the extraction scores.

\noindent\textbf{Reviewer issue recall.} Reviewer issue recall divides matched public review issues by all eligible reference issues. Matching requires the same underlying concern. Mentioning the same topic without the same concern is insufficient.

\noindent\textbf{Aggregation.} Aggregate rates are computed from summed counts. Group percentages are rounded after computation. Gate comparisons retain the same reference claims at every level. The false support axis measures errors among reference issues, while the coverage axis measures recovered reference Supported claims.

\section{Prompt Templates}
\label{app:prompts}

The following templates define the content required at each stage. Runtime inputs provide the manuscript and the evidence available for the current paper at each stage.

\subsection{Claim Extraction}
\label{app:prompt_main}

\begin{promptbox}{Claim extraction}
Extract verifiable claims made by the authors. Retain claims that affect the paper's contribution or the validity of its findings. Split statements whose parts require different evidence. Merge repeated statements expressing the same claim.

For each claim, return its text and source location. Preserve the conditions that determine its scope. Assign a claim type and indicate whether it is central to the paper's contribution.

Identify the evidence needed to assess the claim. Do not infer results absent from the paper. Return structured records with stable claim identifiers.
\end{promptbox}

\subsection{Literature, Theory, and Design Checks}

\begin{promptbox}{Literature verification}
Assess the target claim using the manuscript and retrieved literature. Search for work that could change the judgment. Read the relevant passage before citing a retrieved paper as evidence.

For a novelty claim, compare the proposed contribution with the closest relevant work. Describe the overlap and the remaining difference at the scope claimed by the manuscript.

Link each finding to its source. Indicate whether the evidence supports the claim, establishes a concern, or remains insufficient. Record unresolved questions.
\end{promptbox}

\begin{promptbox}{Theory and experimental design}
Check the reasoning required for the target claim. For a theoretical claim, inspect the assumptions and the relevant derivation. Identify the step that establishes or fails to establish the conclusion.

For an empirical claim, check whether the reported comparison supports the stated conclusion. Examine the evaluation conditions relevant to that claim. Require a controlled comparison when the claim attributes an effect to a particular component.

Return the supporting passage or specific discrepancy. Explain which part of the claim remains unresolved.
\end{promptbox}

\subsection{Code Inspection and Execution}
\label{app:prompt_execution_teaser}

\begin{promptbox}{Code inspection and execution planning}
Trace the target claim to the relevant implementation. Check whether the released code implements the behavior described in the paper.

Prepare an experiment that can resolve the claim within the available budget. Use the documented entry point when possible. Specify the command and expected output. Record the evaluation conditions that must match the paper.

Check the evaluation setup before the main run. Explain how the possible outcomes would affect the claim judgment.
\end{promptbox}

\begin{promptbox}{Execution judgment and repair}
Inspect the saved output of the execution task. Confirm that the intended experiment completed and produced the required result. Compare its conditions with those specified in the plan.

Return pass when aligned evidence supports the claim. Return fail when aligned evidence contradicts it. Return inconclusive when the result cannot resolve the claim. Link the judgment to the supporting output in the logs.

If execution fails, identify the cause from the logs. Propose a repair within the allowed changes. Preserve the scientific implementation and evaluation target. Record the repair before rerunning the same execution task.
\end{promptbox}

\subsection{Claim Audit, Report, and References}
\label{app:prompt_audit}

\begin{promptbox}{Claim audit and report generation}
Check each claim against its cited evidence. Confirm that the evidence addresses the full claim under the stated conditions. Identify unresolved contradictions and missing evidence.

Return the evidence assessment required by the decision rules. Preserve the status assigned by the decision module when generating the report.

For each finding, state the claim and the reason for the judgment. Link to the evidence used. Keep Questioned findings distinct from demonstrated errors. Do not infer an acceptance recommendation.
\end{promptbox}

\phantomsection
\label{app:prompt_refcopilot}
\begin{promptbox}{Reference verification}
Extract references from the bibliography and match them to available bibliographic records. Preserve identifiers supplied by the source.

Report a mismatch only when the retrieved record supports it. If a reference cannot be resolved, state what remains uncertain. Check whether an unresolved item is a legitimate technical report or documentation source before treating it as a citation problem.
\end{promptbox}

\end{document}